\UseRawInputEncoding
\documentclass[prx,twocolumn,superscriptaddress,floatfix,nopacs,nofootinbib,longbibliography]{revtex4-2}

\usepackage{graphicx,amsfonts,amssymb,amsmath,hyperref,enumerate}

\usepackage{adjustbox}
\usepackage{algorithm}
\usepackage{algpseudocode}
\usepackage{pgfplots}
\usepackage{pgf}
\usepackage{lmodern}
\usepackage{import}
\usepackage{xr}
\usepackage{enumitem}
\usepackage{soul}
\usepackage[normalem]{ulem}
\usepackage{multirow}
\UseRawInputEncoding
\usepackage{makecell}

\newif\ifhyper
\hypertrue
\ifhyper
\hypersetup{
   citecolor = {green},
   colorlinks = {true}, 
   urlcolor = {blue} 
}
\fi

\newcommand{\beq}{\begin{equation}}
\newcommand{\eeq}{\end{equation}}
\newcommand{\beqa}{\begin{eqnarray}}
\newcommand{\eeqa}{\end{eqnarray}}
\newcommand{\ket} [1] {\vert #1 \rangle}
\newcommand{\bra} [1] {\langle #1 \vert}

\newcommand{\comment}[1]{}

\def\bra#1{\langle#1\vert}
\def\ket#1{\vert#1\rangle}

\def\Longarrow{\protect\@lra}
\def\@lra{\relbar\joinrel\relbar\joinrel\relbar\joinrel%
          \relbar\joinrel\rightarrow}

\pgfplotsset{compat=1.18}

\begin{document}

\title{Tensor Networks for Explainable Machine Learning in Cybersecurity}

\author{Borja Aizpurua}
\affiliation{Multiverse Computing, Paseo de Miram\'on 170, E-20014 San Sebasti\'an, Spain}
\affiliation{Department of Basic Sciences, Tecnun - University of Navarra, E-20018 San Sebasti\'an, Spain}

\author{Samuel Palmer}
\affiliation{Multiverse Computing, Spadina Ave., Toronto, ON M5T 2C2, Canada}

\author{Rom\'an Or\'us}
\affiliation{Multiverse Computing, Paseo de Miram\'on 170, E-20014 San Sebasti\'an, Spain}
\affiliation{Donostia International Physics Center, Paseo Manuel de Lardizabal 4, E-20018 San Sebasti\'an, Spain}
\affiliation{Ikerbasque Foundation for Science, Maria Diaz de Haro 3, E-48013 Bilbao, Spain}

\begin{abstract} 

In this paper we show how tensor networks help in developing explainability of machine learning algorithms. Specifically, we develop an unsupervised clustering algorithm based on Matrix Product States (MPS) and apply it in the context of a real use-case of adversary-generated threat intelligence. Our investigation proves that MPS rival traditional deep learning models such as autoencoders and GANs in terms of performance, while providing much richer model interpretability. Our approach naturally facilitates the extraction of feature-wise probabilities, Von Neumann Entropy, and mutual information, offering a compelling narrative for classification of anomalies and fostering an unprecedented level of transparency and interpretability, something fundamental to understand the rationale behind artificial intelligence decisions.

\end{abstract}

\maketitle

\section{Introduction}
\label{sec1}

Explainable Artificial Intelligence (XAI) emerges as a cornerstone in the advancement of AI, shedding light on the often opaque intricacies of algorithmic decision-making \cite{expAI}. It strives to render machine learning models that are not only robust and precise but also transparent and comprehensible to human sight. The impetus for XAI is twofold: 1) it cultivates trust and supports robust decision-making by providing clear explanations for outcomes, and 2) it ensures compliance with increasingly stringent transparency regulations. XAI achieves these goals through specific techniques such as LIME (Local Interpretable Model-agnostic Explanations) \cite{lime} and SHAP (SHapley Additive exPlanations) \cite{shap}, where the challenge lies in the tradeoff between model's accuracy and explainability.

Deep learning methods, including neural networks, autoencoders \cite{autoencoder} and GANs \cite{GAN}, excel at detecting intricate data patterns but often act as ``black boxes''. Their complex architectures deliver powerful performance yet hinder the visibility of the decision-making process, posing a challenge to the increasing imperative of interpretability in machine learning. After all, one should be able to explain \emph{why} an algorithm is providing a particular answer, and not a different one. In the search for a solution for this concern, the machine learning community has grown interest in alternative architectures that could make the work.

In this work, we showcase the capabilities of Tensor Networks (TN) \cite{mps_tn} to implement explainable machine learning. While other authors have explored similar trends recently \cite{Su}, we go one step beyond and validate our approach with a real use-case in cybersecurity analytics. 

TNs are a powerful mathematical framework used to represent high-dimensional data efficiently. They factorize vectors and operators in high-dimensional vector spaces into a network of lower-dimensional tensors, enabling to capture complex multi-partite correlations while managing the curse of dimensionality. As such, TNs have a large number of applications in the simulation of complex quantum systems \cite{tn_1}. In the realm of machine learning, TNs have been traditionally considered in two approaches: 1) the tensorization of traditional machine learning  models for enhanced computation, as demonstrated by recent research \cite{tnn}, and 2) the creation of intrinsic tensor network-based models, based on, e.g., Matrix Product States (MPS) \cite{mps_tn, mps_2}. MPS, in particular, stands out as an efficient and transformative approach in explainable AI, displaying not only a high capacity for anomaly detection but also an exceptional degree of interpretability. 

The chosen use-case to test our algorithms comes from cybersecurity. The field of \emph{adversary-generated threat intelligence} focuses on anticipating and understanding the strategies and methodologies employed by cyber-adversaries. By establishing a baseline of normalcy and identifying deviations, MPS can discern intricate patterns of anomalous behavior, often indicative of sophisticated cyber-attacks, while providing insights into the data's underlying structure and dependencies. This demarcation is vital for the early detection and response to potential security incidents, such as network intrusions. While deep learning has proven to be highly effective \cite{nids} in this context, it still lacks interpretability. And, in addition, TNs are known to be more precise than deep learning approaches in anomaly detection for tabular datasets \cite{tn_anomaly_detection}. On top of all this, our paper also shows how TNs, and in particular MPS-based algorithms, are explainable for unsupervised anomaly detection.

This paper is structures as follows: Sec.\ref{sec2} details our MPS methodology and the extraction of interpretative insights from reduced density matrices. Sec.\ref{sec3} presents a comparative performance analysis of MPS in adversary-generated threat intelligence, emphasizing its efficacy in reducing false positives. Sec.\ref{sec4} discusses MPS's interpretability, examining feature probabilities, Von Neumann entropy, and anomaly detection mechanisms. Finally, Sec.\ref{sec5} concludes with a summary of our findings, their impact on explainable AI in cybersecurity, and an outlook on future research directions.

\section{Methodology}
\label{sec2}

This section explores the role of TNs in explainable machine learning, with a focus on MPS as a powerful framework for anomaly detection. We first introduce the general concept of TNs and their relevance to high-dimensional data representations. Then, we delve into the adequacy of MPS within the scope of unsupervised generative modeling for anomaly detection, illustrating its ability to efficiently capture feature correlations. We also explain the mechanisms for extracting valuable insights from MPS, particularly through the use of reduced density matrices, which provide deeper interpretability in machine learning applications.

\subsection{Introduction to Tensor Networks}

Tensor Networks (TNs) are a mathematical framework designed to efficiently represent and manipulate high-dimensional data. They originate from quantum many-body physics and have found applications in machine learning, classical statistical mechanics, quantum chemistry, and even finance \cite{stoudenmire2016supervised, tn_anomaly_detection, PhysRevResearch.4.013006}.

The central idea of TNs is to decompose high-dimensional tensors into a network of lower-dimensional tensors connected by shared indices. This structured representation helps in efficiently capturing correlations in the data, reducing computational complexity, and maintaining interpretability.

Several types of TN architectures exist, each suited for different types of problems \cite{mps_tn, vidal2007entanglement}:
\begin{itemize}
    \item \textbf{Matrix Product States (MPS):} The simplest and most commonly used TN, ideal for one-dimensional sequences, as in anomaly detection.
    \item \textbf{Projected Entangled Pair States (PEPS):} A two-dimensional generalization of MPS, used for more complex correlations but computationally demanding.
    \item \textbf{Multiscale Entanglement Renormalization Ansatz (MERA):} Suitable for hierarchical data and quantum simulations, leveraging scale invariance.
\end{itemize}

In the context of machine learning, TNs enable structured compression of data while preserving essential relationships between features. This ability makes them particularly attractive for interpretable AI applications. Among these TN architectures, MPS stands out as a powerful tool for anomaly detection, efficiently capturing correlations between sequential features while remaining computationally tractable. In the next subsection, we detail the use of MPS in our approach.

\subsection{Unsupervised generative modeling with MPS}
Inspired by the machinery of quantum physics, Ref.\cite{UnsupMPS} introduces a generative model that uses the principles of efficient learning and direct sampling into the fabric of machine learning. At the core of this approach lies the Matrix Product State for the representation of probability distributions over datasets that permits an explicit and easy handling of data features.

Let us briefly sketch the approach in what follows. Given a dataset $\tau \in V = \{0, 1\}^{\otimes N}$, the MPS model represents the probability distribution
\beq
P(v)=\left|\Psi\left(v\right)\right|^{2}
\eeq
where $\Psi(v)$ is the coefficient of the normalized quantum state
\beq
\ket{\Psi} = \sum_{v \in V} \Psi\left(v\right) \ket{v}, 
\eeq
with $v \equiv \left(v_{1}, v_{2},\hdots, v_{N}\right)$. The coefficient of the quantum wave function can be parameterized using an MPS as
\beq
\Psi\left(v_{1},\,v_{2},\,...,\,v_{N}\right)\,=\,{\rm Tr}\left(A^{\left(1\right)v_{1}}\,A^{\left(2\right)v_{2}}\,...A^{\left(N\right)v_{N}}\right),
\eeq
where each $A^{\left(k\right)v_{k}}\,$ is a $D_{k-1} \times D_{k}\,$ matrix.

\subsection{Unsupervised generative modeling with MPS}

Inspired by the machinery of quantum physics, Ref.\cite{UnsupMPS} introduces a generative model that uses the principles of efficient learning and direct sampling into the fabric of machine learning. At the core of this approach lies the Matrix Product State for the representation of probability distributions over datasets that permits an explicit and easy handling of data features. 

\begin{figure}
\centering
\includegraphics[width=1\columnwidth]{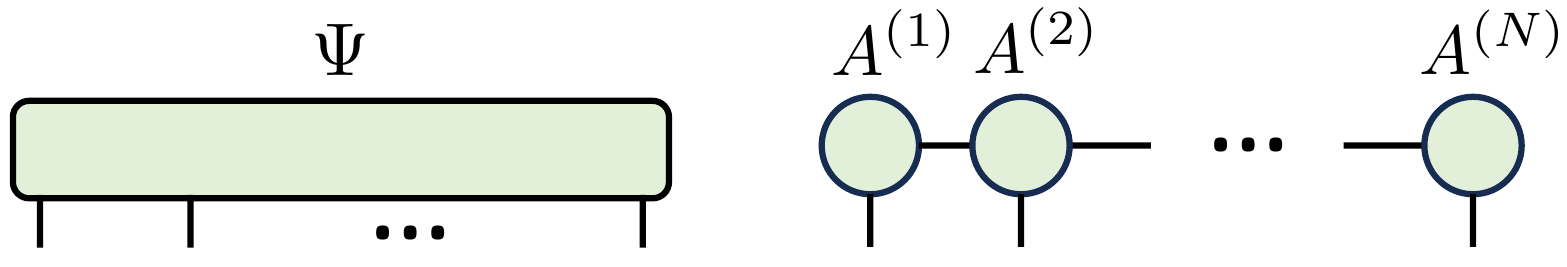}
\caption{[Color online] Representation of a Matrix Product State (MPS) for the coefficient $\Psi(v)$ of quantum state $\ket{\Psi}$, where the system is described by a sequence of interconnected tensors ${A^{(k)}}$. Each block $A^{(k)v_{k}}$ corresponds to a  $D_{k-1} \times D_{k}$ matrix. The vertical lines denote the physical indices $v_{k}$, which represent the state of each subsystem in the quantum state vector $v = (v_{1}, v_{2}, \ldots, v_{N})$, and horizontal lines correspond to bond indices.}
\label{Fig3}
\end{figure}

Let us briefly sketch the approach in what follows. Given a dataset $\tau \in V = \{0, 1\}^{\otimes N}$, the MPS model represents the probability distribution
\beq
P(v)=\left|\Psi\left(v\right)\right|^{2}
\eeq
where $\Psi(v)$ is the coefficient of the normalized quantum state
\beq
\ket{\Psi} = \sum_{v \in V} \Psi\left(v\right) \ket{v}, 
\eeq
with $v \equiv \left(v_{1}, v_{2},\hdots, v_{N}\right)$. The coefficient of the quantum wave function can be parameterized using an MPS as
\beq
\Psi\left(v_{1},\,v_{2},\,...,\,v_{N}\right)\,=\,{\rm Tr}\left(A^{\left(1\right)v_{1}}\,A^{\left(2\right)v_{2}}\,...A^{\left(N\right)v_{N}}\right),
\eeq
where each $A^{\left(k\right)v_{k}}\,$ is a $D_{k-1} \times D_{k}\,$ matrix, as shown in Fig. \ref{Fig3}.

The representational strength of MPS correlates directly with the Von Neumann entanglement entropy of the corresponding quantum state, setting a lower bound for the bond dimensions $D_{k}$ within the system, with larger bond dimensions capturing more complex parameterizations and preserving entanglement. To align the model's probability distribution $P(v)$ with the training data distribution, the model optimizes variationally the parameters of $\Psi(v)$ by minimizing the Negative Log-Likelihood (NLL) function across the training set. The NLL is given by 
\beq
L\,=\,-\,\frac{1}{\left|\tau\right|}\sum_{v\in\tau}^{}\ln\,P\left(v\right),
\eeq
where $\left|\tau\right|$ denotes the size of the training set. 

In the face of the intractability posed by many-body quantum systems, the MPS approach adopts a learning strategy analogous to the Density Matrix Renormalization Group (DMRG) method \cite{dmrg_1, dmrg_2}. DMRG tackles the exponential growth of the Hilbert space by keeping only the most ``relevant'' states of the system and truncating the remainder, thus dynamically adjusting the dimensions of the system based on state importance. Our optimization, akin to DMRG's variational two-site update, dynamically adjusts bond dimensions, ensuring computational resources are allocated efficiently.

The choice of bond dimension D directly affects the computational complexity of the MPS model, scales as $O(N p D^3)$ with $N$ number of features and $p$ physical dimension. In our implementation, we find that increasing D beyond a certain threshold ($\sim$ 150) leads to diminishing returns in anomaly detection performance while significantly increasing training time.

Distinct from traditional generative models, this MPS framework can compute the partition function with linear complexity in the system size and boasts a direct sampling method that constructs samples incrementally. The model allows for deep theoretical understanding of its expressive power, adaptive adjustability, efficient computation of exact gradients and log-likelihood, and efficient direct sampling-position, making it therefore a powerful, efficient and interpretable tool for unsupervised learning and generative tasks.

\subsection{Extracting information from MPS}
\label{sec2c}

The Reduced Density Matrices (RDM) of the MPS contain all the relevant information regarding specific subsystems. As such, these objects allow for a better understanding of the correlations in the model, and are extremely valuable tools to interpret the produced results by the AI system. For a quantum state $\ket{\Psi}$, the RDM $\rho_A$ of subsystem $A$ is defined as 
\beq
\rho_A \equiv {\rm Tr}_{All - A} \left( \ket{\Psi} \bra{\Psi} \right), 
\eeq
with the trace being a \emph{partial trace} over all the degrees of freedom of the system except those of $A$. 

\begin{figure}
\centering
\includegraphics[width=1\columnwidth]{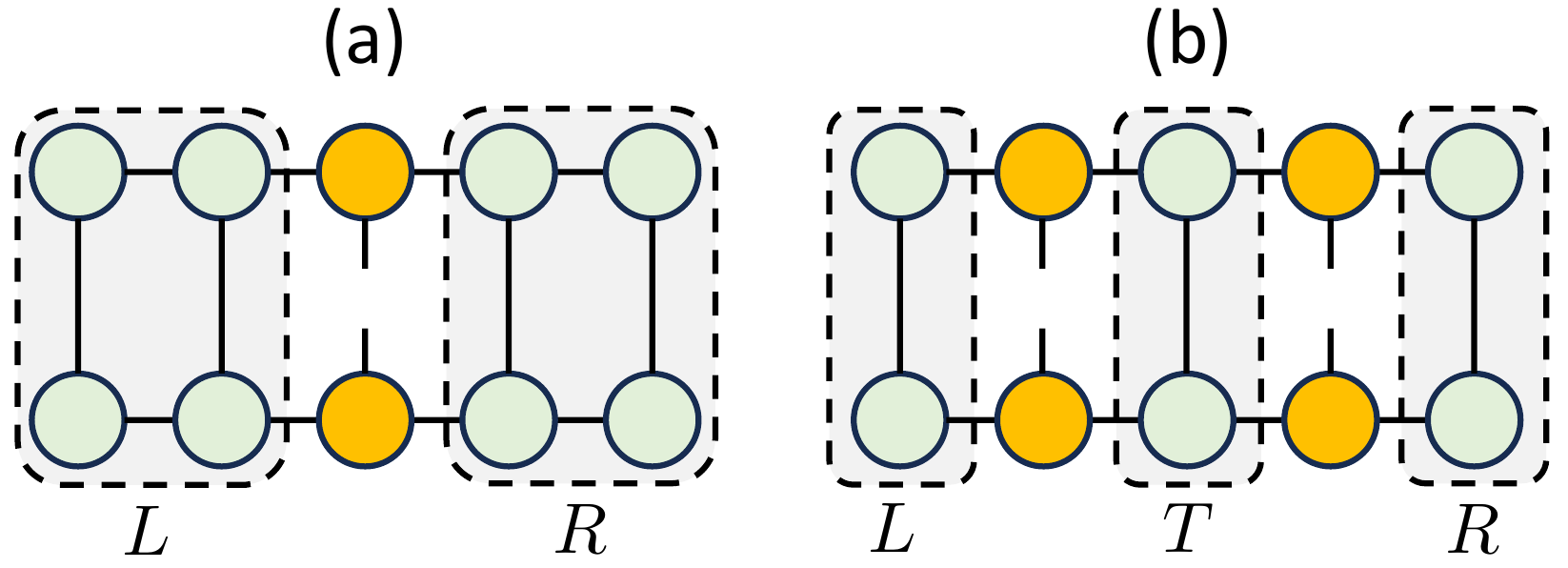}
\caption{[Color online] Graphical representation of the contractions to extract the Reduced Density Matrix (RDM) from an MPS: (a) one site, and (b) two distant sites. The contractions leading to left $L$, right $R$ and central $T$ tensors are also highlighted. Tensors with open physical indices are in orange.}
\label{Fig4}
\end{figure}

To compute RDMs from the MPS, we target specific tensors for a subsystem within state-vector for the complete system. Each tensor, or `site', contributes to state probabilities reflected in the modulus squared of the wavefunction. The RDM of a single site is calculated through a sequence of contractions that progressively build up the left and right environments. This methodically `traces out' the rest of the system to isolate the site's state.  The outcome is a pair of environmental tensors that represent the broader system's influence on the site of interest. The last phase is to contract the target tensor with the environmental tensors, along with its complex conjugate, resulting in the RDM. This matrix not only captures the state of the subsystem but also its entanglements and interactions with the entire system. For subsystems consisting of multiple sites, the RDM can be computed similarly, by including the contraction of intermediate tensors. The computational complexity of these contractions scale polynomially with the largest bond dimension of the MPS, and allow, among other things, the explicit calculation of all possible correlations in the model. In addition, the diagonal entries of the RDM directly convey the state probabilities, providing us with a transparent window into the quantum system's behavior. As a guidance, in Fig.\ref{Fig4} we sketch the calculation of RDMs for one- and two-site subsystems. 

Let us stress that the extracted RDMs are not only rich with information but easily readable right away. This stands in stark contrast to the opaque nature of weights in traditional neural networks, which typically yield little intuitive understanding and require auxiliary methods such as Shapley values for interpretation \cite{shap}. In particular, RDMs enable us to calculate the Von Neumann Entropy, which quantifies the informational content of the considered subsystems. It helps with the evaluation of feature importance, and assess why specific instances are flagged as anomalous through the lens of conditional probabilities and mutual information. 

\section{Adversary-Generated Threat Intelligence}
\label{sec3}

To set the stage for discussing the interpretability capabilities of Matrix Product States (MPS), it is pertinent to first introduce their application within the realm of adversary-generated threat intelligence with a real case scenario in cybersecurity.

For this investigation, we delve into a dataset of cyber-attack vectors obtained from real-life scenarios, provided to us by the cybersecurity company CounterCraft SL. Encompassing a variety of configurations and system states, this dataset mirrors the intricate nuances and complexities that characterize true cyber-attacks. It is against this collection of diverse cyber threats that we later unpack the interpretability brought up by MPS.

More specifically, the dataset deployed in this study is a collection of five distinct subsets, each corresponding to the totality of events recorded across all systems on a specific day. This compilation method offers a longitudinal view of cyber activities, gathering both benign and malicious events. As evidenced by the data, there is a pronounced imbalance between the overall number of events and those identified as part of cyber-attack incidents, which is reflective of the asymmetry typically encountered in real-world cybersecurity datasets. The subsets, named here `CC1' through `CC5', present a diverse landscape of interactions, with `CC1' containing $140,202$ events of which 34 are incident-related, scaling up to `CC5' with $194,766$ events including 822 tied to reported incidents. In total, we have $674,704$ events of which $1,007$ are incident-related. This realistic imbalance poses a unique challenge and underscores the need for a robust anomaly detection system capable of discerning subtle patterns of cyber-threats within predominantly benign data traffic.

To further elucidate the structure and complexity of the dataset, we provide a snapshot of sample entries in Table \ref{tab:sample_entries}. Each entry represents a system event characterized by multiple features, including process paths, usernames, event types, and outcomes:

\begin{table*}[ht]
\centering
\footnotesize 
\caption{Sample entries from the dataset highlighting key features.}
\adjustbox{width=\textwidth}{ 
\begin{tabular}{|p{2.5cm}|p{1.2cm}|p{1.4cm}|p{1.3cm}|p{1cm}|p{1.3cm}|p{1.5cm}|p{1.5cm}|p{1.5cm}|p{1.2cm}|p{1.5cm}|}
\hline
\textbf{Parent Process Path} & \textbf{Process User} & \textbf{Parent Process} & \textbf{Process Name} & \textbf{Real User} & \textbf{Process Path} & \textbf{Type Code} & \textbf{Category} & \textbf{Outcome} & \textbf{Directory} & \textbf{Host IP} \\ \hline
/usr/sbin /nginx & www-data & dash & hostname & root & /usr/bin /sed & Connection Established & process & UNKNOWN & /usr/bin & 10.0.0.4 \\ \hline
/snap/amazon-ssm-agent/6312/ssm-agent-worker & root & ssm-agent-worker & sed & www-data & /usr/sbin /nginx & Create Process & network & SUCCESS & /usr/sbin & 172.31.3.38 \\ \hline
/usr/bin /dash & root & nginx & nginx & root & /usr/bin /hostname & DNSQuery & network & SUCCESS & /usr/bin & 172.31.14.43 \\ \hline
\end{tabular}
}
\label{tab:sample_entries}
\end{table*}

The table highlights critical features such as:
\begin{itemize}
    \item \textbf{Process Information:} Details about the parent and child processes (`Parent Process Path`, `Parent Process`, `Process Path`, `Process Name`).
    \item \textbf{User Details:} The user responsible for the processes (`Process User`, `Real User`).
    \item \textbf{Event Metadata:} Event type (`Type Code`), associated tags (`Tags`), category (`Category`), and outcome (`Outcome`).
    \item \textbf{Network Data:} Host IP address and directories involved.
\end{itemize}

These features encapsulate the breadth of system activities, enabling the detection of cyber-attacks as anomalies. In addition to the core features, an extra Tag feature is derived to classify events based on the MITRE ATT\&CK framework and proprietary classifications such as CounterCraft's unique TTPs (e.g., T9xxx series). Events tagged with specific threat indicators signal potential attacks, such as attack::T9007.002, providing critical insights for identifying malicious activities. The diversity and granularity of these features are essential for the MPS model to effectively discern malicious activities while maintaining interpretability. In the following figures, a total of 23 features are presented. These features were derived by applying numerical encoding and distribution techniques to the original feature set described above, allowing for enhanced representation and facilitating more effective MPS training.

Our methodology for applying the MPS model to anomaly detection in adversary-generated threat intelligence consists of the following steps:

\begin{enumerate}

    \item \textbf{Data Preprocessing:}
    \begin{itemize}
        \item Extract features from logs in various formats (JSON, CSV, PCAP).
        \item Filter out numeric features with zero variance (constant values) and remove duplicated features.
        \item Perform correlation analysis (Spearman and Cram\'er's V) to remove highly correlated numeric features.
        \item If labels are available, analyze feature correlation with labels and select the most important features.
        \item Apply clustering techniques and visualization for exploratory analysis.
        \item Encode categorical features using one-hot encoding or numerical encoding.
        \item Perform time dependency analysis to check for data leakage.
        \item Normalize the dataset and reduce the number of features using Principal Component Analysis (PCA).
    \end{itemize}

    \item \textbf{Feature Encoding:}

    \begin{itemize}

        \item The selected features are numerically encoded to align with the MPS model's input format. This ensures compatibility with the tensor network structure and optimizes the representation of both categorical and continuous data.

    \end{itemize}

    \item \textbf{Training the MPS Model:}
    \begin{itemize}
        \item The MPS model is trained on the first 70\% of benign events, sorted chronologically to ensure a realistic baseline.
        \item The MPS model learns the statistical distribution of normal behavior, capturing feature correlations.
        \item Hyperparameter tuning is performed to optimize performance, balancing computational efficiency and accuracy. The key hyperparameters include:
        \begin{itemize}
            \item \textbf{MaxBondDim}: Maximum bond dimension of the MPS, controlling the model's expressiveness.
            \item \textbf{lr\_shrink}: Learning rate reduction factor if the loss stops improving.
            \item \textbf{loopmax}: Maximum number of training loops.
            \item \textbf{safe\_thres}: Threshold for detecting unstable learning rates.
            \item \textbf{lr\_inf}: Minimum learning rate before training stops.
            \item \textbf{lr\_var}: Initial learning rate.
            \item \textbf{dcs}: Number of descent steps within each optimization loop.
            \item \textbf{cut}: Threshold for truncating small singular values in bond dimension calculations.
        \end{itemize}
        \item These hyperparameters were optimized through a grid search strategy, ensuring that the selected values provided a good balance between performance, stability, and computational cost. In particular, we found that a bond dimension of around 120 offered sufficient expressivity without excessive memory consumption.
    \end{itemize}

    \item \textbf{Anomaly Scoring:}

    \begin{itemize}

        \item For each event in the dataset (including the test data), the MPS model computes a Negative Log Likelihood (NLL) score, representing the likelihood that the event belongs to the baseline normal behavior.

    \end{itemize}

    \item \textbf{Threshold Setting for Anomalies:}

    \begin{itemize}

        \item Anomalies are flagged based on a threshold applied to the NLL scores.

        \item This threshold can be computed using statistical measures such as the mean and standard deviation of NLL values in the training dataset.

    \end{itemize}

    \item \textbf{Evaluation:}

    \begin{itemize}

        \item The remaining 30\% of the data, which includes both benign and attack-related events, is used to evaluate the model.

        \item Performance metrics include detection rates for attack events and false positive rates for benign events.

    \end{itemize}

\end{enumerate}

Our attack-detection system identifies attacks as anomalies. Notably, not all detected anomalies will be attacks. Indeed, there may be low-probability events that are perfectly legitimate (e.g., energy outages, system reboots). Because of this, it is important to adjust the parameters available in the model to minimize the detection of false positives, i.e., anomalies that are not attacks after all. With this in mind, central to the model's success is the implementation of an acceptance threshold for the Negative Log Likelihood (NLL) of each event being generated by the model, so that events that exceed the threshold are classified as anomalous and may be considered by a human analyst to check if they correspond to a real attack. Establishing this threshold is challenging and varies in practice; some analysts may use statistical metrics like the mean and standard deviation to discern attacks within anomalies, while others may opt for the median. Notably, the average NLL of daily benign events is consistently lower than that of attacks, confirming the model's adeptness at flagging attacks as anomalies.

Through rigorous training, which leverages 70\% of the data and respects the chronological order of events, our model effectively captures attack patterns. This temporal fidelity is crucial as shuffling the order of events has been shown to influence model training negatively. The remaining 30\%, containing both benign events and all attack-related events, formed the testing dataset. This approach ensures that the training data simulates realistic scenarios without prior knowledge of future events, aligning with the unsupervised nature of the MPS model. Here, the goal is to detect deviations from the baseline (training data) without explicit attack labels. This setup rigorously evaluates the model's anomaly detection capability on unseen data, including the full spectrum of attack patterns present in the test set.

\begin{figure}
\centering
\includegraphics[width=1\columnwidth]{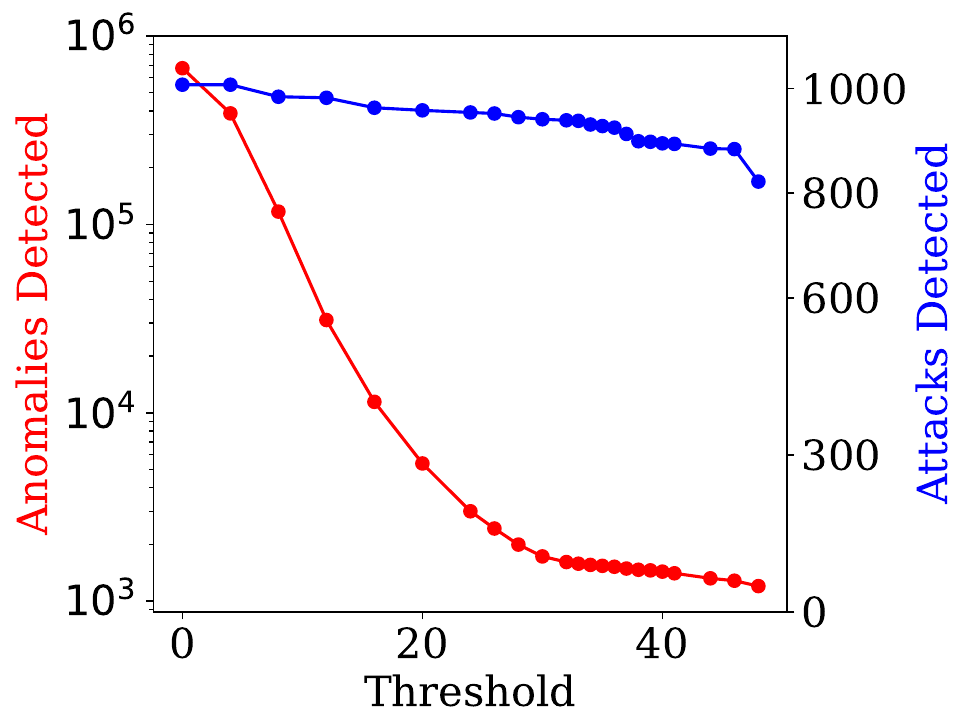}
\caption{[Color online] In red, the number of events categorized as anomalies by the MPS model. In blue, how many of the attacks are among the anomalies detected by the model. The x-axis represents the threshold value for the NLL from which an event is considered an anomaly or not. Notice that both plots have different vertical scales.}
\label{Fig2}
\end{figure}

In Fig.\ref{Fig2} we show one way to determine the threshold of NLL in our model, by analyzing the number of anomalies detected and the number of attack events present in those detected anomalies for multiple threshold values (notice that, in the figure, the scales for both quantities are different). Depending on the specific goal, one can choose a low threshold value and maximize the number of attack events at the expense of having a higher false positive rate, or choose a high value for the threshold and minimize the false positive rate at the expense of not detecting all the attack events inside the detected anomalies. In any case, in the figure we can see that the number of detected anomalies decreases very quickly with the threshold, whereas this is not the case for the number of detected attacks, meaning that our MPS model is remarkably good at pinpointing anomalies that correspond to true attacks.

\begin{figure}
\centering
\includegraphics[width=1\columnwidth]{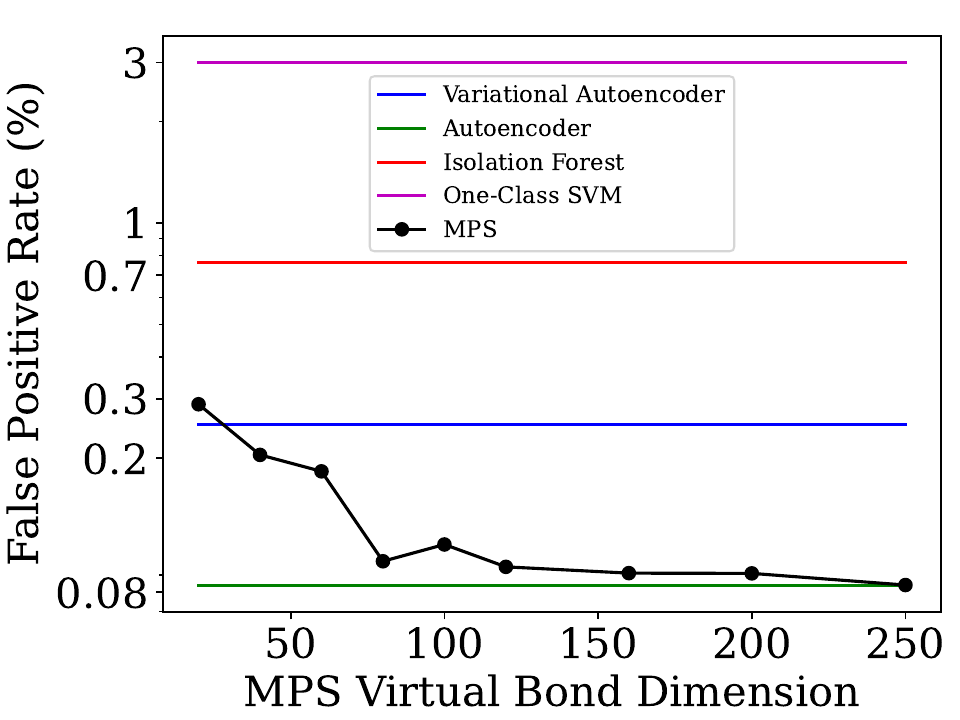}
\caption{[Color online] Comparison of false positive rate amongst traditional machine learning models and the MPS-based unsupervised model. The y-axis is the percentage of false positives while the x-axis is the virtual bond dimension of MPS.}
\label{Fig1}
\end{figure}

The MPS framework does not assume independence among input features, is specifically designed to capture correlations between features through its tensor network structure. Each tensor in the MPS represents a feature, and the connections (bonds) between tensors model the interactions between features.

The bond dimension, denoted as $D$, is a critical parameter that determines the capacity of the MPS to capture feature correlations. A higher bond dimension allows the MPS to represent more complex and stronger correlations among features. Specifically, the expressiveness of the MPS in modeling joint probability distributions increases with the bond dimension, enabling it to capture higher-order dependencies that are common in cybersecurity data.

While a higher bond dimension enables the MPS model to capture richer feature correlations and detect more complex anomalies, it comes with increased computational cost. As mentioned above, the training complexity of an MPS model grows as $O(N p D^3)$. A higher D allows for better expressivity but results in increased memory consumption and longer training times. Therefore, an optimal trade-off is needed: too low a bond dimension may cause underfitting (missing important correlations), whereas too high a bond dimension may lead to overfitting or impractical training times. Our experiments indicate that a bond dimension in the range of 50 to 150 provides a balance between performance and computational efficiency, as evidenced by the decrease in the false positive rate shown in Fig.~\ref{Fig1}. These FPR values are achieved while maintaining a high detection precision of 92.3\%.

In practice, our MPS model achieves remarkable results: we see that by fine-tuning the threshold, an average of 83.5\% of incident-report events are correctly identified as anomalies, and all attack types are detected (100\% attack detection rate). Moreover, the average false positive rate (FPR) stands at a manageable 1.39\%, substantially narrowing the search space for analysts by 98.61\%.

It is worth noting that the time complexity of fitting the MPS model using the DMRG optimization process is higher than that of neural-network-based methods such as autoencoders. This is because we selected a high bond dimension to ensure top performance and interpretable insights, prioritizing explainability and accuracy over training speed. While autoencoders are simpler, faster to train, and effective in anomaly detection, they function as black-box models and lack inherent interpretability. In contrast, the MPS model offers rich insights into feature correlations and anomaly characteristics, making it a better fit for scenarios where interpretability is critical.

Notably, the results could be improved even further by increasing the MPS bond dimension, and/or by using more complex TN structures, able to capture more subtle correlations in the model. Last but not least, our MPS model can be scaled easily to datasets with a huge number of features, and is remarkably memory-efficient, on top of being accurate (as shown) and explainable. This last property is what we discuss in detail in the next section.

\section{MPS interpretability}
\label{sec4}

The fundamental goal of explainable AI is to provide clear explanations of machine learning processes, enabling users to grasp the rationale behind AI-driven decisions, verify the reliability of these systems, and align AI outcomes with ethical and legal standards. TNs such as MPS stand out as a natural candidate for XAI, offering a suite of interpretability capabilities that bridge the division between the abstractness of high-dimensional data and the need for informed decision-making. In the following we discuss the explainability delivered by MPS and its contribution to model's functionality.

\subsection{Direct probability extraction}

As presented in subsection \ref{sec2c}, the RDM's diagonal entries provide a probabilistic view of specific subsystem's state, which is directly connected to the dataset's frequency distribution. The MPS captures and adapts to the correlations in the data during training, allowing security analysts to assess how individual feature distributions contribute to detecting anomalous behaviors. By comparing empirical frequency distributions with MPS-derived probabilities, analysts can pinpoint deviations that may indicate cybersecurity threats. While a simple frequency distribution is akin to a static snapshot, capturing conditional probabilities at a fixed time step, MPS allows for an efficient dynamical representation of other valuable insights, turning exhaustive analysis such as the case of all-to-all feature correlations (scaling exponentially in computational complexity) into a polynomial complexity scaling.

\begin{figure}
\centering
\includegraphics[width=1\columnwidth]{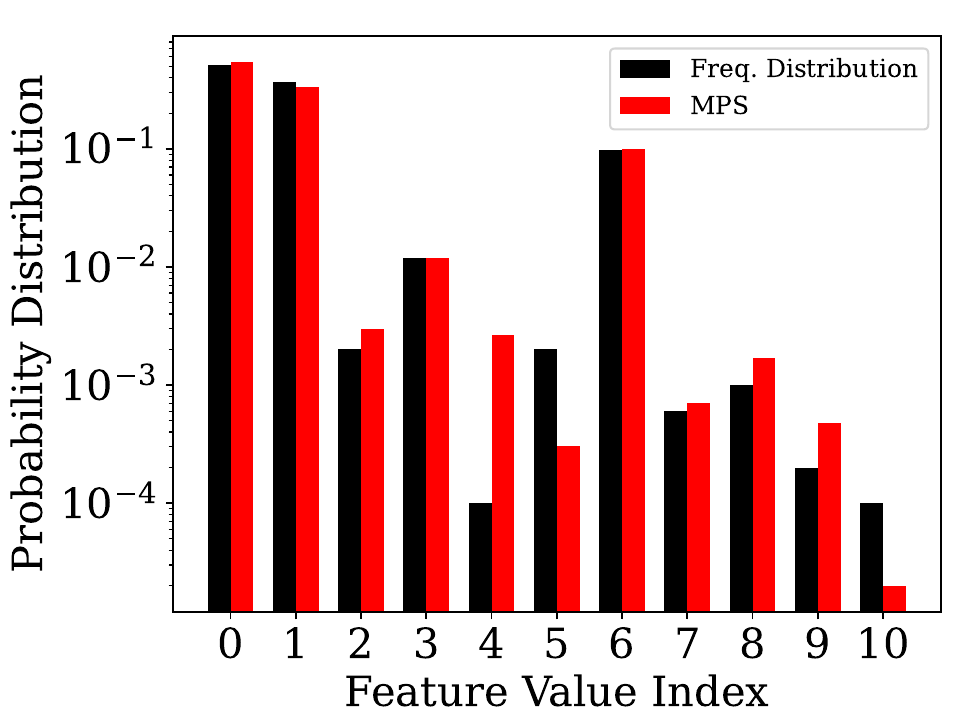}
\caption{[Color online] Comparison of empirical frequency distribution and MPS-derived distribution for a selected feature, represented in a semi-logarithmic scale.}
\label{Fig5}
\end{figure}

In Fig.\ref{Fig5} we illustrate this concept through a semi-logarithmic histogram juxtaposing the empirical frequency distribution of a selected feature against its MPS-derived distribution. As seen in the plot, both quantities are similar but not equivalent, due to the influence of correlations between features. In stark contrast to classical deep learning models, where such direct extraction of probabilities is not inherently possible, MPS offers a unique advantage here. Classical models require the use of external tools to approximate probability distributions, often through additional post-processing steps. MPS, however, integrates this capability into the core of its architecture. 

\subsection{Von Neumann entropy}

In quantum-mechanical terms, the Von Neumann Entropy measures the degree of entanglement between two systems. In particular, the entropy of a subsystem $A$ and the rest of the system is defined as 
\beq
S \equiv - {\rm Tr}\left( \rho_A \log \rho_A\right), 
\eeq
being $\rho_A$ the RDM of subsystem $A$. In our setting, the entropy of a single MPS site measures the degree of entanglement between the feature corresponding to that site and the rest of the features in the system. In machine learning jargon, this is a measure of the contextual interdependencies of the features in the model. 

The concept of Von Neumann Entropy is analogous to the Shannon entropy in classical information theory \cite{shannon} but adapted for quantum systems and reduced density matrices. In the end, the Von Neuamn entropy of a RDM is nothing but the Shannon entropy of its eigenvalues. It quantifies the correlation between a subsystem with the rest of the system, thus providing insight into how independent a feature is within the broader data context. This metric plays a crucial role in cybersecurity anomaly detection, as it helps distinguish between features that contribute independently to an event and those that are contextually bound to cyber threats. Features with high entropy often indicate complex dependencies, such as coordinated attacks or lateral movements within a compromised network, while low-entropy features may correspond to isolated system events. This allows to differentiate between what might be deemed normal and anomalous within a dataset, effectively utilizing ``context" to pinpoint deviations.

\begin{figure}
\centering
\includegraphics[width=1\columnwidth]{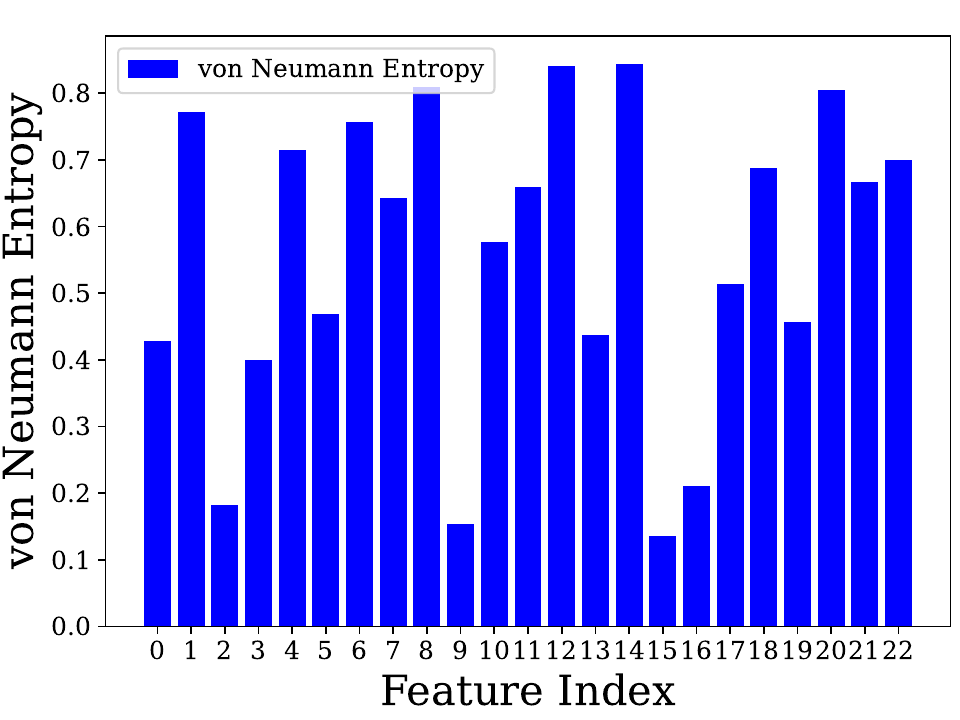}
\caption{[Color online] Von Neumann entropy of single MPS sites across various features in the dataset. Features with higher entropy values signify greater entanglement with the rest of the system, indicating a high degree of contextual interdependency.}
\label{Fig6}
\end{figure}

In Fig.\ref{Fig6}, we present the Von Neumann entropy calculated for each feature within our dataset. It is evident from the plot that certain features exhibit higher entropy values, e.g., features 8, 12, 14, and 20. Such high values of the entropy are indicative of a substantial degree of correlation with other features. Conversely, features 2, 9, and 15 have entropies less than 0.2, suggesting low correlation with other features and therefore hig degree of independence. 

\begin{figure}
\centering
\includegraphics[width=1\columnwidth]{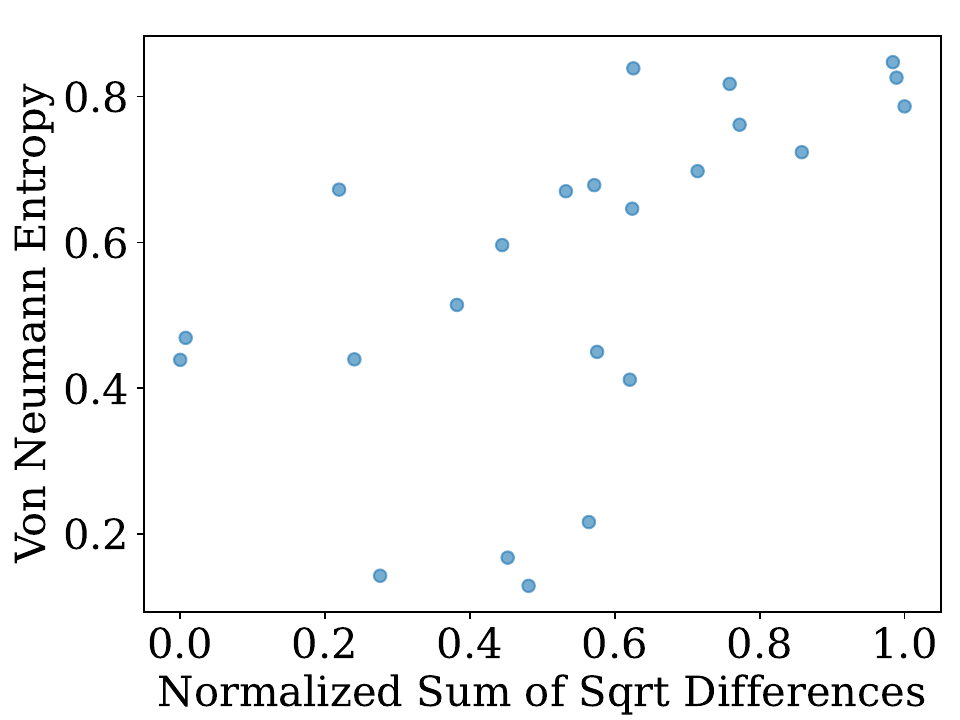}
\caption{[Color online] Normalized sum of square root differences between frequency and MPS distributions plotted against Von Neumann entropy for all features.}
\label{Fig7}
\end{figure}

To further explore the relationship between a feature's entanglement with the rest of the system and its distribution representation, we show Fig.\ref{Fig7}. In this figure, we plot the normalized sum of the square root differences between the frequency and MPS distributions over all features against their Von Neumann entropies. While not universally consistent, a pattern emerges across most features: those with higher entanglement often exhibit a more significant disparity between their MPS distribution and the empirical frequency distribution. This was to be expected: the more entangled a feature is with the rest, the less independent is its probability distribution, and therefore the less similar to a simple frequency analysis. 

The capacity to directly derive such insights from an MPS model is what sets it apart from classical deep learning approaches. Where classical models might require supplementary techniques such as activation maximization or feature visualization to approximate these insights, MPS inherently provides them. 

\subsection{Feature importance}

The assessment of feature importance is a crucial aspect of interpretability in machine learning. It involves quantifying the relative impact of each feature within a dataset on the model's output. Within the MPS framework, this evaluation is conducted through a detailed analysis of the probability distributions derived from the MPS for each feature. These distributions allow for the calculation of average probabilities for the possible values of each feature across all instances, resulting in a mean probability that reflects the feature's general tendency.

To further analyze the differences between benign and attack events, we computed the mean probabilities of feature values for both benign and attack instances as given by the MPS model. In our model, the likelihood of each event $\mathbf{v} = (v_1, v_2, \dots, v_N)$ is computed as the squared magnitude of the wavefunction $\Psi(\mathbf{v})$, which, for an MPS model, can be expressed as a product of probabilities associated with each feature:

\begin{equation}
P(\mathbf{v}) = |\Psi(\mathbf{v})|^2 = \prod_{i=1}^{N} P_i(v_i)
\end{equation}

where $P_i(v_i)$ is the probability associated with the $i$-th feature value $v_i$ derived from the local density matrices in the MPS model. The overall likelihood of an event is thus the product of the probabilities of its feature values. We compute the mean of these overall likelihoods separately for benign events and attack events to compare their distributions. Additionally, the Negative Log Likelihood (NLL) for an event is calculated as:

\begin{equation}
\mathrm{NLL} = -\log P(\mathbf{v}) = -\sum_{i=1}^{N} \log P_i(v_i).
\end{equation}

This NLL represents how unlikely an event is under the model trained on benign data; higher NLL values indicate greater anomalies.

\begin{table}
\centering
\small
\begin{tabular}{|c|c|c|}
\hline
\textbf{Feature} & \makecell{\textbf{Mean Prob.}\\ \textbf{ (Benign)}} & \makecell{\textbf{Mean Prob.}\\ \textbf{ (Attack)}}\\ \hline
0 & 0.41 & 0.47 \\
1 & 0.18 & 0.08 \\
2 & 0.87 & 0.03 \\
3 & 0.55 & 0.02 \\
4 & 0.26 & 0.02 \\
5 & 0.41 & 0.48 \\
\vdots & \vdots & \vdots \\
19 & 0.5 & 0.61 \\
20 & 0.16 & 0.24 \\
21 & 0.31 & 0.41 \\
22 & 0.31 & 0.47 \\
\hline 
 \makecell{\textbf{Mean Overall}\\ \textbf{Likelihood}} & 2.14e-10 & 6.06e-18 \\ \hline
\textbf{Mean NLL} & 9.669 & 17.217 \\ \hline
\end{tabular}
\caption{Mean probabilities of feature values and overall likelihoods for benign and attack instances.}
\label{feat_imp}
\end{table}

Table \ref{feat_imp} presents the computed mean probabilities for features in benign and attack contexts, which can be instrumental in feature selection. Features with high probabilities in the benign category and low probabilities in the attack category, such as `data.parent\_process\_path`, `data.process\_user\_name`, and `data.parent\_process\_basename` (features 2, 3, and 4), are indicative of normal operational patterns. Their high likelihood under benign conditions and rarity during attacks signify their importance in identifying abnormal behavior. These features represent critical aspects of process-level activity: the execution path of a parent process, the user responsible for the process, and the process's basename. Their behavior aligns with real-world expectations, where deviations in these parameters often signal potential malicious activities, such as unauthorized user access or unusual process execution paths. Conversely, features with a higher mean in the attack category than in the benign may be less informative for detecting anomalies, as their higher occurrence in attack scenarios does not contribute to distinguishing attacks from benign behavior. In addition, the total probability values, obtained by the product of probabilities across all features, reveal a striking disparity: the probability associated with attack features is several orders of magnitude lower than that of benign features, and the mean NLL for attack events is significantly higher. This stark contrast explains why the MPS model effectively discriminates between benign operations and cyberattacks: the improbability of the conjunction of attack features underscores their anomaly.

In traditional deep learning approaches, determining feature importance would typically involve a suite of ad-hoc techniques such as Sensitivity Analysis, Gradient-based Techniques, or Feature Attribution Methods like SHAP \cite{shap}. These methods, while powerful, require additional computational steps and can sometimes yield results that are not immediately interpretable.

\subsection{Anomaly identification}

One of the most compelling aspects of using Matrix Product States (MPS) in anomaly detection is the model's ability to provide clear explanations for the identification of individual data instances as anomalies. This granularity is especially valuable in cybersecurity, where understanding the specific reason behind an anomaly alert is crucial for effective response.

When the MPS model flags a row as anomalous, the underlying reason can be investigated by examining the individual probabilities of each feature's values within that instance. The MPS allows for an inspection of these probabilities, which are readily available from the model's parameters. Low probabilities associated with certain feature values contribute to the instance being deemed anomalous, as they deviate from the model's learned representation of normal behavior. Conversely, feature values with high probabilities are in alignment with the expected patterns and therefore do not contribute to the anomalous classification.

To quantify the anomaly, one can calculate the product of the individual feature probabilities for the row in question. By applying the negative log-likelihood (NLL) to this product, we obtain a scalar value representing the degree of anomaly. If this value surpasses a predefined threshold, the instance is classified as an anomaly.

The ability to perform this analysis is inherent to the MPS framework and does not require additional auxiliary tools. This direct approach contrasts sharply with methods employed in conjunction with traditional deep learning models, such as autoencoders. While such models can effectively identify anomalies, they do not inherently provide insights into why a particular instance is classified as such. Unpacking the decision-making process typically involves more complex procedures, such as sensitivity analysis or activation maximization, which attempt to approximate the model's reasoning ad-hoc.

\begin{figure}
\centering
\includegraphics[width=0.5\columnwidth]{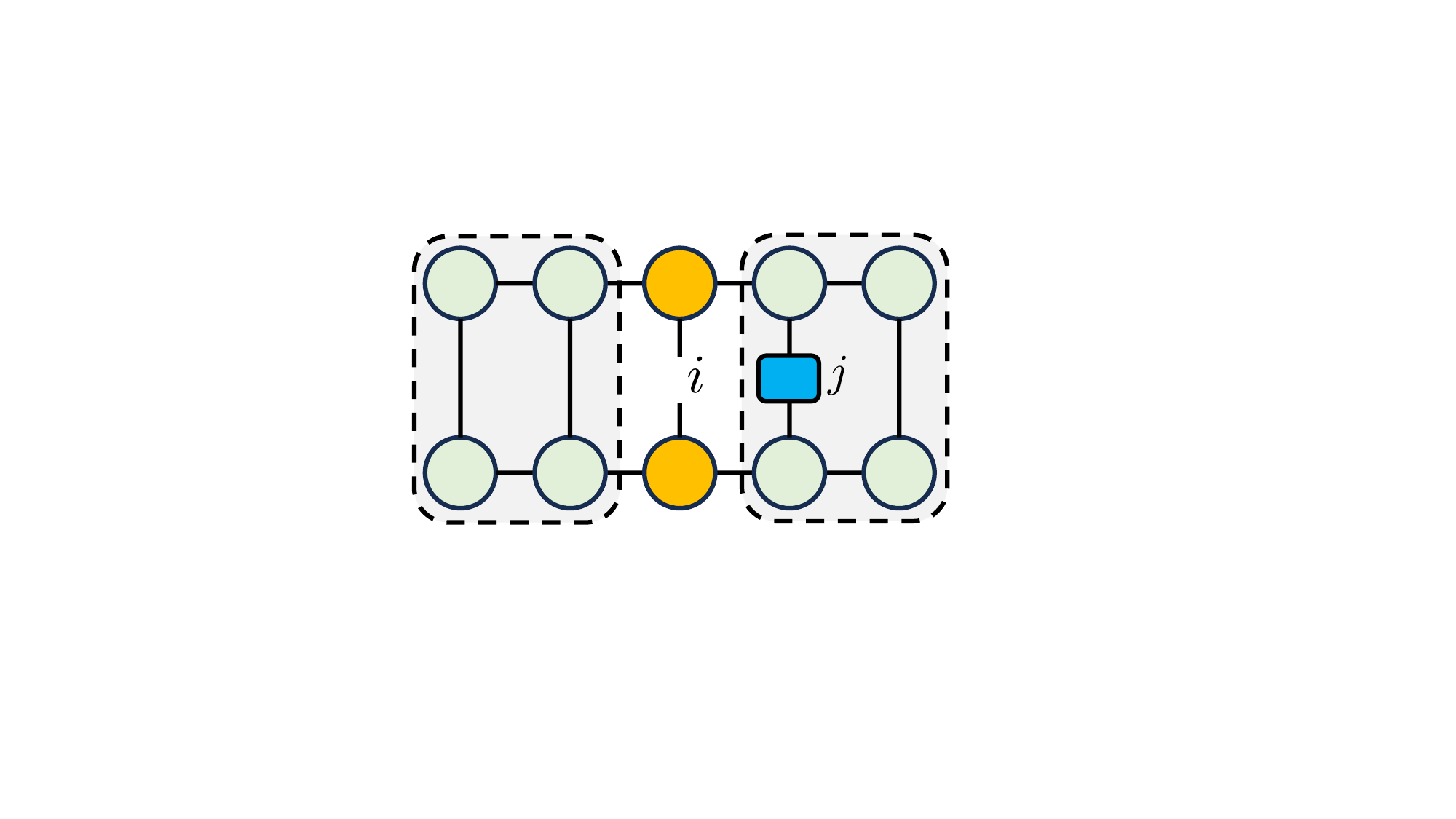}
\caption{[Color online] TN contraction needed to compute the conditional  RDM for one site of an MPS, highlighting how the state of one feature can affect the state distribution of another when its value is fixed. The operator inserted in the physical index of site $j$ is diagonal and selects the value for that feature, acting as a filter. the contractions leading to left and right tensors are highlighted, and tensors with open indices are in orange.}
\label{Fig8}
\end{figure}

Through the MPS model, not only can we specify which features and their respective values are driving the anomalous classification, but we also gain a quantitative measure of the anomaly's severity. A human can then interpret \emph{why an anomaly is an anomaly}, and take action correspondingly. This interpretability is crucial, transforming anomaly detection from a mere alerting mechanism into a diagnostic tool that offers actionable insights, which is a substantial advancement in the field of XAI.

\subsection{Conditional Probabilities}

Conditional probabilities within the MPS framework provide a nuanced understanding of how the state of one feature influences the state of another. By fixing the value of the physical index of one tensor, we can observe the resultant changes in the reduced density matrix (RDM) of another tensor, effectively capturing the conditional relationships within the data.

For the sake of illustration, consider the process of obtaining a conditional density matrix where we select tensor $i$ from the MPS to derive its RDM while fixing the physical index of tensor $j$ at value $k$. This operation is akin to asking the question: ``What is the state of feature $i$ given that feature $j$ is known to be $k$?'' The methodology involves a sequence of contractions similar to those used in obtaining an unconditional RDM, but inserting an additional operator at site $j$ that selects value $k$ for that physical index (in practice, a diagonal matrix at site $j$  with a ``1" at index value $k$ and ``0" elsewhere). A similar strategy can be applied to evaluate more complex conditional probabilities, as required. Fig. \ref{Fig8} graphically represents the tensor network contraction for the conditional RDM from the MPS.

In our practical case, an example can be seen when fixing feature 21 to value ``0" and observing its effect on feature 1. Table \ref{cond_prob} showcases the probabilities of feature 1 both when feature 21 is not conditioned and when it is conditioned to be ``0".

\begin{table}
\centering
\begin{tabular}{|c|c|c|}
\hline
~\textbf{Value Index}~ & ~\textbf{Not Conditioned}~ & ~\textbf{Conditioned} ~\\
\hline
\hline
0  & 0.54     & 0.78    \\
1  & 0.33     & 0.15    \\
2  & 0.003    & 0.004   \\
3  & 0.012    & 0.036   \\
4  & 0.00267  & 0.003   \\
5  & 0.000303 & 0.003   \\
6  & 0.099    & 0.009   \\
7  & 0.0007   & 0.0008  \\
8  & 0.0017   & 0.002   \\
9  & 0.00048  & 0.0007  \\
10 & 0.00002  & 0.00003 \\
\hline
\end{tabular}
\caption{Comparison of the probabilities for feature 1 with and without conditioning on feature 21 being ``0". The conditioned probabilities reveal how the state of feature 21 influences the probability distribution of feature 1.}
\label{cond_prob}
\end{table}

In traditional deep learning, obtaining such conditional probabilities is not as direct. Autoencoders, for example, can be used to infer conditional probabilities by leveraging latent space representations in a supervised setting. However, this requires additional steps beyond the standard training of the autoencoder.

\subsection{Mutual Information}

Mutual information, a concept deeply rooted in information theory, quantifies the amount of information obtained about one random variable from another. In the context of MPS, this metric can reveal the total correlation between two features, shedding light on their interdependence within the system. For a pair of subsystems or features within an MPS, mutual information can elucidate the extent of their correlation, beyond the scope of linear relationships captured by traditional correlation coefficients. This is particularly valuable when considering the complex, non-linear interactions prevalent in high-dimensional data, as often encountered in cybersecurity and other advanced analytics domains.

The mutual information $I(X;Y)$ between two random variables $X$ and $Y$ is defined as
\begin{equation}
I(X; Y) \equiv \sum_{y \in Y} \sum_{x \in X} p(x, y) \log \left(\frac{p(x, y)}{p(x)p(y)}\right),
\end{equation}
where $p(x,y)$ is the joint probability distribution function of $X$ and $Y$, and $p(x)$ and $p(y)$ are the marginal probability distribution functions of $X$ and $Y$, respectively. In this case, taking into account that the von Neumann entropy of RDM $\rho_A$ is defined as $S \equiv -\text{Tr}(\rho_{A}\log \rho_{A})$, the mutual information can be rewritten as
\begin{equation}
I(A;B) = S(\rho_{A}) + S(\rho_{B}) - S(\rho_{AB}).
\end{equation}

\begin{figure}
\centering
\includegraphics[width=1\columnwidth]{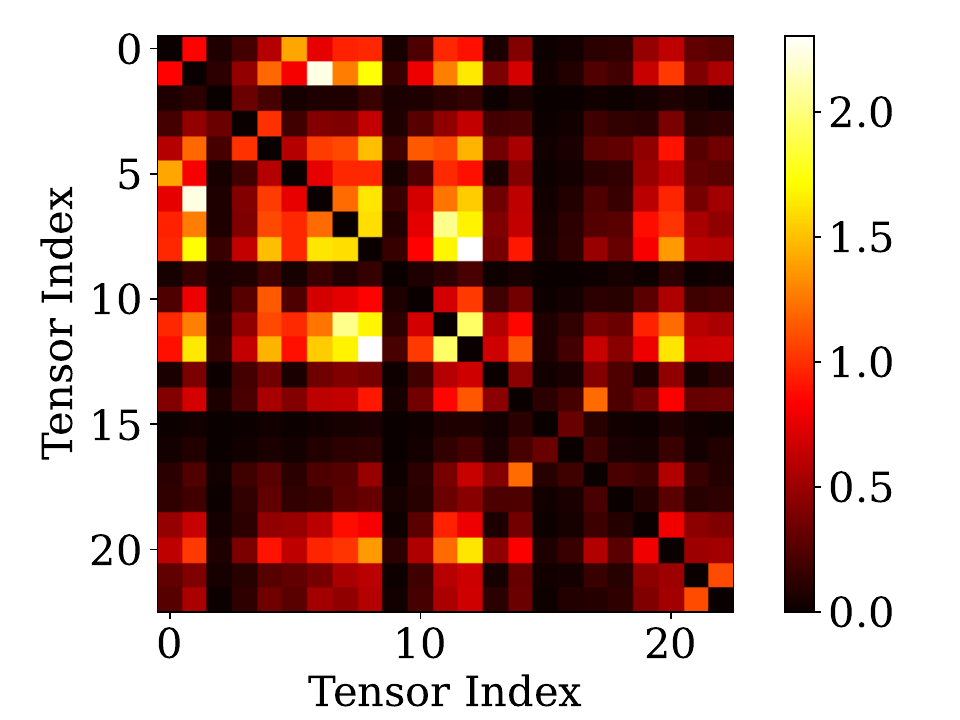}
\caption{[Color online] Mutual information heatmap for the different features. We can observe that some pairs of features are highly correlated while others are very independent.}
\label{Fig10}
\end{figure}

The incorporation of mutual information into our MPS framework allows to understand the intricate correlations between features within the system. The mutual information heatmap in Fig.\ref{Fig10} is particularly valuable in cybersecurity, as it reveals which features share the most information, highlighting patterns such as correlated network events or coordinated attacker behavior. By clustering highly correlated features, security teams can optimize intrusion detection rules and reduce redundant monitoring of less informative variables. Beyond interpretability, this metric holds practical utility in optimizing the MPS model. By rearranging columns to position highly correlated features adjacent to each other, we can refine the efficiency of the virtual bond dimension, ensuring a more resource-effective and potent representation of the data's entangled nature. In addition, those pairs of features with large mutual information could even be clustered together, therefore reducing the complexity of the problem. 

In classical deep learning models, the analogous insights are generally not as straightforward to obtain. While certain techniques can estimate mutual information or capture feature interdependencies (such as Canonical Correlation Analysis (CCA) \cite{canonical} or neural estimation approaches), they often do not provide the direct and intrinsic measure that an MPS model with mutual information capabilities would.

\bigskip 

\section{Conclusions and Next Steps}
\label{sec5}

In this paper we have discussed in detail the potential of TN methods to develop explainable and interpretable machine learning algorithms. We have shown the validity of the approach for an MPS algorithm designed for clustering, with the application to a real-life use case in cybersecurity. Our research has unveiled the potential of MPS as a powerful tool for unsupervised anomaly detection, particularly within the domain of adversary-generated threat intelligence. The key findings demonstrate that MPS not only rivals but in certain respects surpasses the performance of traditional machine learning models such as autoencoders and GANs, especially in terms of explainability and interpretability. The application of MPS to a cybersecurity dataset revealed its robustness in identifying cyberthreats with a low false positive rate, a critical parameter in cybersecurity operations. The ability of MPS to provide direct probability extraction, assess feature importance, and elucidate on the reasons behind the classification of data points as anomalies underscores its utility in operational environments where understanding the `why' behind a model's decision is as vital as the decision itself. In addition, the research has highlighted the interpretability capabilities of MPS, offering insights into Von Neumann entropy, feature importance, conditional probabilities, and mutual information directly from the model's parameters without the need for additional tools or techniques. These capabilities have profound implications for explainable AI, setting a new benchmark for transparency in machine learning applications. In addition, the methods presented in this paper allow the capability to generate synthetic data in order to improve training of complex models (due to lack of data availability) and activity generation for deception systems. This would also help in the development of behavior-based models, which are better against unknown attacks than rule-based models.

The methods developed in this work can be improved in many ways. For instance, we could consider more advanced TN structures beyond MPS, with more intricate patterns of correlations. Furthermore, as part of future work, we aim to validate our MPS-based approach on additional well-established cybersecurity datasets to assess its robustness across different data distributions. Specifically, we plan to apply our model to:
\begin{itemize}
    \item \textbf{KDD Cup 1999} \cite{KDDCUP1999}: One of the most widely used intrusion detection datasets, containing simulated attacks in a network traffic environment.
    \item \textbf{CIC-IDS2017} \cite{CIC-2017}: A modern intrusion detection dataset with real-world traffic and attack types collected by the Canadian Institute for Cybersecurity.
    \item \textbf{UNSW-NB15} \cite{UNSW-NB15}: A dataset capturing hybrid real and synthetic network traffic to simulate real-world cyber threats.
\end{itemize}
These datasets contain a diverse set of attack types and real-world traffic patterns, making them ideal benchmarks for evaluating the scalability and robustness of our method. This future validation will allow us to assess whether our model retains its effectiveness across varying data distributions and attack scenarios. We leave this as an important direction for future research.

\bigskip 

{\bf Acknowledgements.-} We acknowledge Donostia International Physics Center (DIPC), Ikerbasque, Basque Government, Diputaci\'on de Gipuzkoa, European Innovation Council (EIC) and Tecnun for constant support, as well as insightful discussions with the teams from Multiverse Computing, DIPC and Tecnun on the algorithms and technical implementations. Special thanks to the team at CounterCraft SL, for providing the motivation for this work,  the dataset and numerous technical discussions. This work was supported by the Diputaci\'on Foral de Gipuzkoa through the ``Q4DecOps: Algoritmo cu\'antico para detecci\'on de anomal\'ias en entornos de ciber-enga\~{n}o'' project (2022-QUAN-000026-01).

\bigskip 
{\bf Data availability statement:} authors are open to reasonable requests.

\bibliography{biblio2.bib}
\end{document}